\pdfoutput=1

\documentclass[11pt]{article}

\usepackage[]{acl}

\usepackage{times}
\usepackage{url}
\usepackage{latexsym}
\usepackage{amsmath,amssymb,amsfonts}
\usepackage{algorithmic}
\usepackage{graphicx}
\usepackage{textcomp}
\usepackage{kotex}
\usepackage{booktabs}
\usepackage{multirow}
\usepackage{arydshln}
\usepackage{tabularx}
\usepackage{color}
\usepackage{amssymb}
\usepackage{mathtools}
\usepackage{makecell}
\usepackage{hyperref}
\usepackage{appendix}
\usepackage{caption}
\usepackage{todonotes}

\usepackage[T1]{fontenc}
\usepackage[utf8]{inputenc}
\usepackage{microtype}

\usepackage{multirow}
\usepackage{multicol}
\usepackage{float}
\usepackage{vcell}
\usepackage{booktabs}
\usepackage{makecell}
\usepackage[normalem]{ulem}
\useunder{\uline}{\ul}{}
\usepackage{rotating}
\usepackage{lipsum}
\usepackage{cite}

\usepackage{epstopdf}
\usepackage{cell}
\usepackage[utf8]{inputenc}

\usepackage{xstring}
\usepackage{xspace}

\usepackage{color}
\usepackage{kotex}  

\newcommand{\datasetname}{\textsc{KoCHET}\xspace}

\title{\datasetname: a Korean Cultural Heritage corpus for Entity-related Tasks\\ \vspace*{.5\baselineskip}}

\author{Gyeongmin Kim\thanks{~~These authors have equally contributed to this work}, Jinsung Kim\footnotemark[1], Junyoung Son\footnotemark[1], Heuiseok Lim\thanks{~~Corresponding author} \\
Korea University, Korea \\
\texttt{\{totoro4007, jin62304, s0ny, limhseok\}@korea.ac.kr}
}

\begin{document}
\maketitle
\begin{abstract}
As digitized traditional cultural heritage documents have rapidly increased, resulting in an increased need for preservation and management, practical recognition of entities and typification of their classes has become essential.
To achieve this, we propose \datasetname - a Korean cultural heritage corpus for the typical entity-related tasks, i.e., named entity recognition (NER), relation extraction (RE), and entity typing (ET).
Advised by cultural heritage experts based on the data construction guidelines of government-affiliated organizations, \datasetname consists of respectively 112,362, 38,765, 113,198 examples for NER, RE, and ET tasks, covering all entity types related to Korean cultural heritage.
Moreover, unlike the existing public corpora, 
modified redistribution can be allowed both domestic and foreign researchers.
Our experimental results make the practical usability of \datasetname more valuable in terms of cultural heritage.
We also provide practical insights of \datasetname in terms of statistical and linguistic analysis.
Our corpus is freely available at \href{https://github.com/Gyeongmin47/KoCHET}{https://github.com/Gyeongmin47/KoCHET}.
\end{abstract}


\section{Introduction}\label{intro}
Recently there has been an increasing interest in the preservation of national historical artifacts and traditional cultural heritage, and also grows up the importance of effective management of them through digitization and archival.
As the amount of digitized information materials increases rapidly, information extraction (IE) tasks in natural language processing (NLP), such as named entity recognition (NER), relation extraction (RE), and entity typing (ET), have become an essential and fundamental step in the field of historical document analysis.

Despite the necessity of a well-refined entity-centric corpus specialized in domestic cultural heritage, unfortunately, there no exists any cultural heritage domain-specialized corpus in Korean.
Moreover, conventional entity-related systems deal only with a coarse set of entity types such as person, location, and organization which is significantly limited in terms of application.
This absence of cultural heritage domain-specialized corpus and narrow coverage of entity types hinders the effective digitization of domestic historical documents because training the model with general corpus for entity-related tasks cannot afford to learn enough significant entity types such as pagodas, historical sites and intangible heritage, and their relations.
Furthermore, not in the cultural heritage domain, the existing entity-related datasets supervised by the public institutions have a complicated procedure for data acquisition, and they are also restricted from modification and redistribution.
These cumbersome procedures and restrictions have been stumbling blocks for researchers against the rapid increase in digitized cultural heritage materials over the past few decades.

To address these difficulties against the conservation of Korean cultural heritage, we introduce a new dataset collection called \datasetname - \textbf{Ko}rean \textbf{C}ultural \textbf{H}eritage corpus for \textbf{E}ntity-related \textbf{T}asks, a high-quality Korean cultural heritage domain-specialized dataset for NER, RE, and ET tasks.
For corpus construction, we crawled the e-museum digitized data of the National Museum of Korea\footnote{\href{https://www.emuseum.go.kr/main}{https://www.emuseum.go.kr/}} (including data from all 50 museums) as the source text which is for the interested public.
We selectively used resources from the museums in which the details of artifacts were registered; moreover, for the completeness of the attribute data, we limited the chronological range of the data from the prehistoric era to the Korean Empire era, excluding the Japanese colonial period.
For the annotation, the categorization for classes and attributes appropriate was defined and developed following the \textit{2020 Named Entity Corpus Research Analysis}\footnote{\href{https://www.korean.go.kr/front/reportData/reportDataView.do?mn_id=207&report_seq=1050&pageIndex=1}{https://www.korean.go.kr}} which was published under the guidelines as institutional organizations.

As our corpus focuses on the entity features, it has more detailed and abundant entity types including diverse cultural heritage artifacts, compared to the existing accessible datasets that aim to deal with several downstream tasks in addition to entity-related tasks.
Furthermore, the ET of \datasetname is the first freely available corpus for the ET task in Korea.
In addition to providing these values, this paper provides detailed statistics and linguistic analysis of \datasetname for each entity-related task to demonstrate their applicability and enhance understanding of the data, along with baseline experiments with language models.

Our contributions are summarized as follows:

\begin{itemize}
    \item We introduce \datasetname designed for entity-related tasks. This guarantees a high-quality corpus without restrictions regarding modification and redistribution. Moreover, to the best of our knowledge, the ET corpus is the first proposed corpus in Korean.
    \item We categorized the detailed entity types specialized in the cultural heritage domain, which is essential for preserving our cultural and historical artifacts, thereby contributing as an alternative to the increased demand for the digitalized archiving of cultural heritage documents.
    \item We prove the applicability of our entity-abundant corpus in each task by providing statistics and linguistic analysis, along with the experiments with pre-trained language models.
\end{itemize}

\section{Related Works}
As domains that require expertise, such as the cultural heritage, contain entities or relationships that rarely appear in general domains, the necessity of a corpus specialized in the domain is obvious.
Despite such demand, Korean does not yet have a corpus specialized in the cultural heritage area, unlike other languages.
\subsection{General cultural heritage corpora}
There have been the disclosures of corpora in an effort to preserve traditional culture including the cultural heritage, composing data from the perspective of the entity-related tasks that we deal with.
For example, these include a Czech NER corpus constructed based on public optical character recognition data of Czech historical newspapers~\citep{hubkova2020czech}, a Chinese corpus suitable for the computational analysis of historical lexicon and semantic change~\citep{zinin-xu-2020-corpus}, and an English corpus that is one of the most commonly used large corpora in diachronic studies in English~\citep{alatrash2020ccoha}.

\subsection{Korean public corpora}


\paragraph{The National Institute of Korean Language}, which is an institution that has established the norms for Korean linguistics, constructed a large-scale dataset\footnote{\href{https://stdict.korean.go.kr/}{https://stdict.korean.go.kr/}} for the study of new computational linguistics of Korean \citep{kim2006korean}.
\paragraph{AI HUB} is a massive dataset integration platform\footnote{\href{https://aihub.or.kr/}{https://aihub.or.kr/}} hosted by the National Information Society Agency (NIA)\footnote{\href{https://www.nia.or.kr/}{https://www.nia.or.kr/}}, a government-affiliated organization. 
To support the development of the Korean artificial intelligence industry for the NLP field, the NIA disclosed domain-specific corpora and 27 datasets have been released or are being prepared.
\paragraph{Electronics and Telecommunications Research Institute}, as part of the Exo-brain project\footnote{\href{http://exobrain.kr/pages/ko/result/outputs.jsp}{http://exobrain.kr/pages/ko/result/outputs.jsp}}, provides corpora for NLP tasks such as morphological analysis, entity recognition, dependency parsing, and question answering, and guidelines for building such high-quality corpora\footnote{\href{https://www.etri.re.kr/}{https://www.etri.re.kr/}}.
In addition to public datasets opened by public institutions, there is a Korean dataset publicly available for free without the requirement for an access request.
\paragraph{Korean Language Understanding Evaluation (KLUE)} dataset was recently released to evaluate the ability of Korean models to understand natural languages with eight diverse and typical tasks~\citep{park2021klue}. 
The tasks include natural language inference, semantic textual similarity, dependency parsing, NER, and RE.

\section{\datasetname}
Following the guidelines of Korean institutional organizations, \datasetname is a domain specialized corpus for cultural heritage, which ensures quality and can be freely accessed. 
In this section, we report the annotation process and guidelines in detail.


\subsection{Annotation Process} 
To improve the quality of annotations on our entity-rich corpus related to cultural heritage, we conducted the annotation process based on expertise in the cultural heritage domain. 

\paragraph{Annotation Guidelines}
The raw corpus annotated by each annotator is equally divided by the category.
The annotators were instructed to follow two types of rules by the aforementioned entity guidelines in Section~\ref{intro}; one is related to tagging units and categories, and the other is the principle of unique tagging.
The minimum unit is based on one word for the tagging units and categories. In addition, it is applied only to cases written in Korean, where the notation is possible. It is not tagged in the case of Chinese characters and English, but if it is read in Korean, it is included in the tagging range.
For the principle of unique tagging, there are cases of duplication in entities that belong to two or more semantic regions. This guideline grants a single tag to a semantically suitable word and refers to assigning only one tag by prioritizing it accordingly.
There are two cases in which this principle should be applied. The first case is where the entity belongs to two semantic categories regardless of the context. The second refers to the case where it may vary depending on the context. In both cases, tagging is determined according to the pre-defined priority.

\paragraph{Annotator Training and Cross-Checking}
We recruited 34 college and graduate annotators who have been professionally educated on the cultural heritage domain in Korea to participate in the annotation process. 
All annotators were trained for a week, and each of them was familiarized with the annotation guideline and conducted practice annotation on test samples.
The annotation team met once every week to review and discuss each member’s work during the annotation process. 
All entity types and relations were reviewed
by four cross-checking annotators, afterward, were additionally checked by two expert supervisors.
The discrepancy between annotators on the annotated entity types and relations is also discussed and agreed upon in the period.
These procedures allowed the reliability and validity of \datasetname on the cultural heritage objects to be improved.

\subsection{Schema for Task Annotation}
\subsubsection{Named Entity Recognition}\label{subsec:NER}

\begin{table}[h]
\centering
\scalebox{0.675}{
\label{table:ner_label}
\begin{tabular}{l|r|r|r} 
\toprule[1.5pt]
\multicolumn{1}{c|}{\multirow{2}{*}{\textbf{Label}}}                   & \multicolumn{1}{c|}{\textbf{Train}} & \multicolumn{1}{c|}{\textbf{Dev}} & \multicolumn{1}{c}{\textbf{Test}}  \\ 
\cmidrule{2-4}
\multicolumn{1}{c|}{}                                                  & \multicolumn{3}{c}{\textbf{Counts (\%)}}                                                                     \\ 
\midrule
Artifacts (AF)                                                         & 91,453 (35.57)                      & 11,374 (35.54)                    & 11,366 (35.35)                     \\
Person (PS)                                                            & 51,758 (20.13)                      & 6,455 (20.17)                     & 6,744 (20.97)                      \\
Term (TM)                                                              & 25,781 (10.02)                      & 3,175 (9.92)                      & 3,159 (9.82)                       \\
Date (DT)                                                              & 23,636 (9.19)                       & 2,943 (9.20)                      & 3,078 (9.57)                       \\
\begin{tabular}[c]{@{}l@{}}Political \\ location (LCP)\end{tabular}    & 20,076 (7.80)                       & 2,375 (7.42)                      & 2,384 (7.41)                       \\
Civilization (CV)                                                      & 15,404 (5.99)                       & 1,929 (6.03)                      & 1,835 (5.71)                       \\
Material (MT)                                                          & 8,893 (3.45)                        & 1,160 (3.62)                      & 1,046 (3.25)                       \\
Location (LC)                                                          & 6,881 (2.67)                        & 857 (2.68)                        & 881 (2.74)                         \\
Animal (AM)                                                            & 4,376 (1.70)                        & 578 (1.81)                        & 566 (1.76)                         \\
Plant (PT)                                                             & 3,952 (1.53)                        & 549 (1.72)                        & 498 (1.55)                         \\
\begin{tabular}[c]{@{}l@{}}Geographical \\ location (LCG)\end{tabular} & 2,821 (1.09)                        & 354 (1.11)                        & 348 (1.08)                         \\
Event (EV)                                                             & 2,045 (0.79)                        & 254 (0.79)                        & 248 (0.77)                         \\
\bottomrule[1.5pt]
\end{tabular}
}
\caption{\label{table:ner_label}The counts of entities and their distributions (\%) in our NER data.}
\end{table}

As described in Table \ref{table:ner_label}, we defined 12 entity types. They were tagged with the character-level beginning-inside-outside (BIO) tagging scheme, which is the generally adopted method for sequence labeling problems.
For example, ``아시아 (Asia): Geographical Location (LCG)'' is tagged as ``아: B-LCG,'' ``시: I-LCG,'' ``아: I-LCG.'' Therefore, we evaluated the model not only with entity-level F1 score but also with character-level F1 score \citep{park2021klue}.
\paragraph{Label Description}
\begin{itemize}
    \item \textbf{Artifacts (AF)} generally refer to objects created by humans corresponding to common and proper nouns and also include cultural properties. Therefore, artificial materials such as buildings, civil engineering constructions, playground names, apartments, and bridges fall under this category.
    \item \textbf{Person (PS)} is a category for content related to people, including real persons, mythical figures, fictional characters in games/novels, occupations, and human relationships.
    \item \textbf{Term (TM)} includes the color, direction, shape, or form that describes an artifact. 
    Patterns and drawings are classified as TM, owing to the characteristics of movable cultural properties.
    \item \textbf{Civilization (CV)} is defined as terms related to civilization/culture. It targets words classified by detailed civilizations/cultures, such as clothing and food.
    \item \textbf{Date (DT)} includes all entities related to date and time, such as date, period, specific day, or season, month, year, era/dynasty. However, in the case of an unclear period that cannot be tagged with a separate entity, tagging is not performed.
    \item \textbf{Material (MT)} includes a substance used as a material or an expression for the substance. 
    In other words, it indicates the entity corresponding to the detailed classification of a substance (metal, rock, wood, etc.).
    When an entity can be tagged as both natural objects (AM, PT) and MT, tagging as MT takes precedence.
    \item \textbf{Geographical location (LCG)}, \textbf{Political location (LCP)}, and \textbf{Location (LC)} are defined as geographical names, administrative districts, and other places, respectively.
    \item \textbf{Animal (AM)} and \textbf{Plant (PT)} are defined as animals and plants, respectively, excluding humans. If it is applied as a subject of a picture, it is also included in the category of animals and plants. 
    \item \textbf{Event (EV)} contains entities for a specific event/accident. In principle, social movements and declarations, wars, revolutions, events, festivals, etc., fall under this category and should be classified only if they exist as a separate entity.

\end{itemize}

\subsubsection{Relation Extraction}
Unlike the other existing corpora, our corpus has the advantage of capturing various relationships between multiple entities that are included in a sentence because more than one relation can exist per raw sentence.
We consider the relations between annotated entities in the NER annotation procedure.
In the case of certain tokens, it can be a subject or an object depending on the relationship with other tokens. A relationship in the form of a self-relationship between identical tokens does not exist.
\begin{table}[t]
\centering
\scalebox{0.65}{
\label{table*:relabel}
\begin{tabular}{l|rrr} 
\toprule[1.5pt]
\multicolumn{1}{c|}{\multirow{2}{*}{\textbf{Label}}} & \multicolumn{1}{c|}{\textbf{Train}} & \multicolumn{1}{c|}{\textbf{Dev}} & \multicolumn{1}{c}{\textbf{Test}}  \\ 
\cmidrule{2-4}
\multicolumn{1}{c|}{}                                & \multicolumn{3}{c}{\textbf{Counts (\%)}}                                                                     \\ 
\midrule
A depicts B                                          & 14,157 (22.09)                      & 1,803 (22.45)                     & 1,711 (21.85)                      \\
A documents B                                        & 10,214 (15.94)                      & 1,244 (15.49)                     & 1,220 (15.58)                      \\
A hasSection B                                       & 6,542 (10.21)                       & 818 (10.19)                       & 776 (9.91)                         \\
A servedAs B                                         & 6,546 (10.22)                       & 780 (9.71)                        & 740 (9.45)                         \\
A hasCreated B                                       & 6,136 (9.58)                        & 759 (9.45)                        & 744 (9.50)                         \\
A OriginatedIn B                                     & 5,456 (8.51)                        & 679 (8.45)                        & 663 (8.47)                         \\
A consistsOf B                                       & 4,331 (6.76)                        & 569 (7.09)                        & 586 (7.48)                         \\
A isConnectedWith B                                  & 3,489 (5.44)                        & 501 (6.24)                        & 461 (5.89)                         \\
A fallsWithin B                                      & 3,454 (5.39)                        & 415 (5.17)                        & 483 (6.17)                         \\
A isUsedIn B                                         & 1,906 (2.97)                        & 238 (2.96)                        & 244 (3.12)                         \\
A hasTime B                                          & 934 (1.46)                          & 111 (1.38)                        & 95 (1.21)                          \\
A wears B                                            & 798 (1.25)                          & 97 (1.21)                         & 86 (1.10)                          \\
A hasCarriedOut B                                    & 112 (0.17)                          & 15 (0.19)                         & 19 (0.24)                          \\
A hasDestroyed B                                     & 5 (0.01)                            & 2 (0.02)                          & 3 (0.04)                           \\
\bottomrule[1.5pt]
\end{tabular}
}
\caption{\label{table*:relabel}Relation counts and distributions (\%) for our RE corpus.}
\end{table}
\begin{figure*}[t]
	\centering
	\includegraphics[width=1\linewidth]{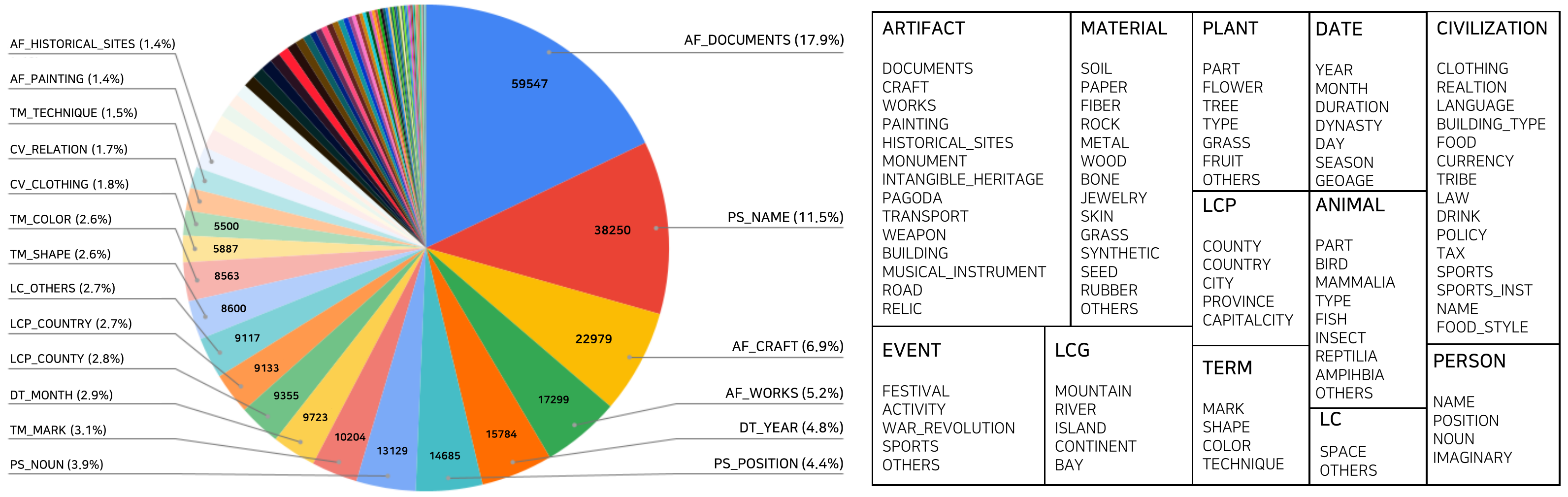}
	\caption{Visualization of all the labels that cover 84\% of the entity types is shown on the left side, and 106 general and fine-grained entities with their distributions (\%) are shown on the right side.}
	\label{fig:cropped_distribution}
\end{figure*}
As shown in Table \ref{table*:relabel}, our RE corpus consists of 14 labels, and these were defined based on the Encyves ontology research of the National Culture Research Institute\footnote{\href{http://dh.aks.ac.kr/Encyves/wiki/index.php?title=\%EB\%AF\%BC\%EC\%A1\%B1\%EA\%B8\%B0\%EB\%A1\%9D\%ED\%99\%94_\%EC\%98\%A8\%ED\%86\%A8\%EB\%A1\%9C\%EC\%A7\%80&oldid=99197}{http://dh.aks.ac.kr/Encyves/wiki}}.
\paragraph{Label Description}
\begin{itemize}
    \item ``\textbf{A depicts B}'' implies the relationship between an object and its color, shape or pattern, etc. For example, ``Green Door'' corresponds to this relationship. It can also represent a descriptive relationship such as ``Picture of a place-the place where it was taken'' or ``Picture of a person-the person who is the object of the painting.''
    \item ``\textbf{A documents B}'' implies ``{\raise.17ex\hbox{$\scriptstyle\sim$}} records -.'' ;a relationship such as ``Record-The person who records it'' can be represented by this. It also indicates the relationship like a record written on an object such as ``Postcard-Explanation" or a specific language written on a document such as ``Record-Chinese characters.''
    \item ``\textbf{A hasSection B}'' indicates ``{\raise.17ex\hbox{$\scriptstyle\sim$}} is located at -.'' It represents the relationship between a statue, building, or specific attraction and a location, such as a certain city and place. 
    \item ``\textbf{A servedAs B}" implies ``{\raise.17ex\hbox{$\scriptstyle\sim$}} is the role of -,'' which corresponds to the relationship between a person, and his/her position or occupation, etc.
    \item ``\textbf{A hasCreated B}'' demonstrates, for example, ``Person-Documents'' or ``Person-Painting,'' which refers to the relationship between a person and a document such as a book, map, or drawing, or his/her activities to record works.
    \item ``\textbf{A OriginatedIn B}'' means ``{\raise.17ex\hbox{$\scriptstyle\sim$}} is discovered at –'' or ``{\raise.17ex\hbox{$\scriptstyle\sim$}} is produced at -(time).'' It indicates that cultural property is produced at a specific time such as ``Craft-Year" or is discovered at a particular place such as ``Object-Place," or is produced at a certain site such as ``Document-Place."
    For example, the relation between earrings and tombs or a newspaper and the company of the newspaper fall into this.
    \item ``\textbf{A consistsOf B}'' refers to the relation between an object and its raw ingredients, such as soil, iron, and wood that constitute an object.
    \item ``\textbf{A isConnectedWith B}'' represents a person-to-person association. The relationships between two positions or a person and the position he or she holds do not fall into this.
    \item ``\textbf{A fallsWithin B}'' implies ``{\raise.17ex\hbox{$\scriptstyle\sim$}} is denominated as -.'' It indicates the relationship of alternate names such as ``Person-Specific name," or between a name and designation in front of the name, or between words that refer to synonymous concepts such as ``Verse-Poetry.''
    \item ``\textbf{A isUsedIn B}'' indicates ``{\raise.17ex\hbox{$\scriptstyle\sim$}} is used for the purpose of -'' or literally ``{\raise.17ex\hbox{$\scriptstyle\sim$}} is used in -.'' For example, it can also indicate the material used for a certain object, such as ``Raw material-Clothes.'' The relationship between an object and the place where the object is used, such as a signboard and a palace, or the relationship between certain means of performing a function and an object such as ``Bowl-Rice cake'' can correspond to this category.
    \item ``\textbf{A hasTime B}'' implies ``{\raise.17ex\hbox{$\scriptstyle\sim$}} has happened at -.'' For example, it can indicate the relationship between a particular event and a specific date, such as ``Presidential election-1928." The relation between a specific date and a certain work, such as the year of production of a work and the year of construction of a building, can fall under this category, for example, ``Year-Craftwork."
    \item ``\textbf{A wears B}'' implies ``{\raise.17ex\hbox{$\scriptstyle\sim$}} puts - on.'' For instance, not only clothes such as school uniforms but also crafts, etc. may correspond to the object argument. 
    \item ``\textbf{A hasCarriedOut B}'' indicates ``- is caused by {\raise.17ex\hbox{$\scriptstyle\sim$}}.'' It can represent a relationship between a specific organization or group and an event conducted by it, such as a festival or social movement.
    \item ``\textbf{A hasDestroyed B}'' implies the event that caused destruction such as ``War-Destroyed place," or the collapse of a country in a specific year such as ``Country-Year,'' or the relationship in which a building, structure, monument, etc. is destroyed at a particular period.
\end{itemize}

\begin{table*}[t]
    \centering
    \scalebox{0.8}{
    \begin{tabular}{l|l} 
    \toprule[1.5pt]
    \textbf{Sentence with Entity Mention}           & \textbf{Entity Types}                                                  \\ 
    \midrule
    \begin{tabular}[c]{@{}l@{}}\textbf{조선시대}에는 전통 관습을 잇기 위한 많은 향로가 제작되었다. \\(In the \textbf{Joseon dynasty}, many fragrance burners were created \\for traditional customs.)\end{tabular}  & \begin{tabular}[c]{@{}l@{}}\textcolor{blue}{\textbf{DT\_DYNASTY}},~DT\_DURATION\\LCP\_COUNTRY,~LCP\_CITY,~LCP\_COUNTY\\LC\_OTHERS,~AF\_DOCUMENTS\end{tabular}   \\ 
    \midrule
    \begin{tabular}[c]{@{}l@{}}노란 바탕의 \textbf{모란}이 양쪽에 그려져 있다. \\(The yellow background \textbf{peony} is drawn on both sides.)\end{tabular} & \begin{tabular}[c]{@{}l@{}}\textcolor{blue}{\textbf{PT\_FLOWER}},~PT\_TYPE,~PT\_OTHERS,\\TM\_SHAPE \end{tabular}                                                \\
    \midrule
    \begin{tabular}[c]{@{}l@{}}\textbf{19세기 후반}~청주의 재정을 파악할 수 있는 자료가 있다.\\(There are data to comprehend the finances of Cheongju in\\ the \textbf{late 19th century}.)\end{tabular}            & \begin{tabular}[c]{@{}l@{}}\textcolor{blue}{\textbf{DT\_YEAR}},~DT\_DYNASTY,~DT\_DURATION \end{tabular}                                                                                                         \\
    \bottomrule[1.5pt]
    \end{tabular}
    }
    \caption{\label{table:example_et}Examples including entity mentions and their fine-grained entity types. Entity mentions and the correct types in the given context are bold. All fine-grained entity types are shown in Figure~\ref{fig:cropped_distribution}.}
\end{table*}

\subsubsection{Fine-grained Entity Typing}\label{section:ET}

Given a sentence and entity mention within it, the ET task predicts a set of noun phrases that describe the mention type.
For example, in ``김홍도는 조선 후기의 화가이다. (\textit{Kim Hong-do was a painter of the Joseon era of Korea.}),'' \textit{Joseon} should be typed as ``dynasty/Date'' and not ``country/Location.''
This typification is crucial for context-sensitive tasks such as RE, coreference resolution, and question answering (e.g., ``In which era was Kim Hong-do, an artist?''). 
Unlike high resource languages, we found that the Korean corpus for the ET task has not been released. In dealing with this data scarcity problem and promoting universal studies, we release a Korean ET task corpus for the first time, to the best of our knowledge.

The schema for the ET task was designed with reference to the data construction process of the Fine-Grained Entity Recognition dataset~\citep{ling2012fine}.
Considering the properties of the cultural heritage domain, we categorized the 12 general entity types aforementioned in the NER task (Section \ref{subsec:NER}) into a fine-grained set of 94 types with detailed meanings.
Particularly, the cultural taxonomy defined in the \textit{Cultural Properties Protection Law}\footnote{\href{www.cha.go.kr}{www.cha.go.kr}} was applied to AF, and the 2004 Cavalier-Smith's classification system~\citep{cavalier2004only} was applied to the biological scope of PT and AM.
All fine-grained entity types are detailed in Figure \ref{fig:cropped_distribution}.

The fine-grained entities for entity-related downstream tasks in the cultural heritage domain enable a more detailed contextualized representation for each entity mention than the previous typing schemas, which only predict relatively coarse types of entities.
Table \ref{table:example_et} lists three example sentences with entity mention that can represent several fine-grained types.
Given a sentence with an entity mention, the appropriate type that describes the role of the entity span in the sentence should be predicted.
Our fine-grained entity types can embrace all the existing general types and categorize them in greater detail. Accordingly, they can let models understand richly the noun phrases including entity, compared to when the models are trained to predict only relatively coarse types.
For Figure \ref{fig:cropped_distribution}, the circle on the left shows the visualization of fine-grained entity types that possess approximately 84\% among all labels in the corpus, and the set on the right shows the detailed distributions of all fine-grained types.
Each example includes 2.94 fine-grained entities on average; there are up to nine several fine-grained entity types per entity. 
The category to which the most entities belong is ``AF\_DOCUMENTS," which possesses 17.9\%, and that on the second place is ``PS\_NAME," having 16.7\%.\newline

\paragraph{Label Description}

\begin{itemize}
    \item \textbf{12 general} types: PS, AF, AM, CV, DT, EV, PT, MT, TM, LC, LCG, LCP
    \item \textbf{94 fine-grained} types, which were mapped to the cultural heritage-specialized fine-grained entity labels, were inspired by prior works~\citep{ling2012fine, gillick2014context, choi2018ultra}.
\end{itemize}

\begin{table*}[h]
\centering
\scalebox{0.825}{
\centering
\begin{tabular}{c| p{15cm}}
\toprule[1.5pt]
\textbf{Index} & \textbf{Example sentences}\\ \midrule
1 & 앞면 좌측 하단에 `한번사신레꼬-드는승질상밧고거-나믈느지는안슴니다' 문구가 있음.  \\ \cdashline{2-2}[0.8pt/1.2pt]
 & There is a phrase ‘한번사신레꼬-드는승질상밧꼬거-나믈느지는안슴니다’(archaic Korean) in the left corner of the front side. \\ \midrule
2 & 1면에는 안창호씨(安昌浩氏)의 연설, 편집실 여언(餘言) 등의 기사가, $\cdot\cdot\cdot$, 인쇄됨. \\ \cdashline{2-2}[0.8pt/1.2pt]
 & On the first page, articles such as Mr. Changho Ahn(安昌浩氏)(Chinese character)’s speech and editorial comments(餘言)(Chinese character), $\cdot\cdot\cdot$, were printed.
\\ \midrule
3 & `戰爭の訓示', $\cdot\cdot\cdot$, 등의 기사와  일본 언어학자 가나자와 쇼자부로(金澤庄三郞, 1872$\sim$1967)의 현대 국어 음운에 대한 연구물인 「朝鮮語發音篇」의 일부를 게재함. \\ \cdashline{2-2}[0.8pt/1.2pt]
 & `戰爭の訓示(Japanese)', $\cdot\cdot\cdot$, the articles and 「朝鮮語發音篇」(Chinese character), the part of a study on the modern Korean phonology of Japanese linguist Kanazawa Shouzaburou(金澤庄三郎(Chinese character), 1872∼1967) were published.
\\ 
\bottomrule[1.5pt]
\end{tabular}
}
\caption{Example sentences contained in our corpus. These examples include not only Korean but also Japanese and Chinese characters. Also, they contain archaic expressions that are not used in modern times. These characteristics make it more suitable for the learning of cultural heritage domain. Note that we omitted some of the words in the sentence for brevity.\label{table:corpus-examples}}
\end{table*}

\subsection{Analysis on \datasetname}
\subsubsection{Diachronic and Linguistic Analysis}
There are mainly two differences between the entities in the proposed corpus and those commonly used.

First, archaic expressions that are not used in modern times are frequently shown in our corpus.
Specifically, such expressions continually appear when ancient documents or historical artifacts are quoted.
Let us consider the phrase ``한번사신레꼬-드는승질상밧고거-나믈느지는안슴니다'' in sentence 1 in Table \ref{table:corpus-examples}.
Although it is written using syllables of modern Korean, the grammar and the vocabulary are fairly dissimilar from those of contemporary Korean, such as word spacing and syllabification, i.e., separation rule between the units of the word.
When translating the sentence with quotation marks into modern Korean, it can be expressed as ``한번 사신 레코드는 성질상 바꾸거나 무르지는 않습니다 (Once a record is purchased, it cannot be exchanged or refunded due to its characteristics)."

Second, several entities contained in \datasetname written in Korean are followed by the descriptions written in either Chinese or Japanese characters.
For example, as shown in sentence 2 in Table \ref{table:corpus-examples}, the description with Chinese characters in parentheses follows the entity ``안창호씨," and is usually written such as ``안창호씨(安昌浩氏)."
Further, Japanese characters are also present throughout the corpus, enhancing the polyglot property of the corpus, as shown in sentence 3.
Therefore, to fully understand such expression types in our corpus, multilingual factors of language models should be considered; particularly in the case of token classification tasks, in which the meaning of each token directly affects the model performance.

\begin{table}[t]
\centering
\scalebox{0.825}{
\begin{tabular}{c|l|c|c|c} 
\toprule[1.5pt]
\multicolumn{2}{c|}{\begin{tabular}[c]{@{}c@{}}\textbf{Task}\\\end{tabular}} & \textbf{Train} & \textbf{Dev} & \textbf{Test}  \\ 
\midrule
\multirow{2}{*}{\textbf{NER}} & \# of examples                               & 89,884         & 11,245       & 11,233         \\
                              & \# of entities                               & 393,076        & 32,003       & 32,153         \\ 
\midrule
\multirow{2}{*}{\textbf{RE}}  & \# of examples                               & 31,012         & 3,876        & 3,877          \\
                              & \# of relations                              & 64,080         & 8,031        & 7,831          \\ 
\midrule
\multirow{2}{*}{\textbf{ET}}  & \# of examples                               & 90,558         & 11,320       & 11,320         \\
                              & \# of mentions                               & 266,209        & 33,226       & 33,395         \\
\bottomrule[1.5pt]
\end{tabular}
}
\caption{\label{table:total-statistics}Statistics of \datasetname for each task.}
\end{table}

\begin{table*}[t]
\centering
\scalebox{0.85}{
\centering
\begin{tabular}{l|cc|cc}
\toprule[1.5pt]
\multicolumn{1}{c|}{\multirow{2}{*}{\textbf{Model}}} & \multicolumn{2}{c|}{\textbf{NER}} & \multicolumn{1}{c|}{\textbf{RE}} & \multicolumn{1}{c}{\textbf{ET}} \\ \cmidrule{2-5}
& \multicolumn{1}{c|}{Entity F1 ($\sigma$)} & \multicolumn{1}{c|}{Character F1 ($\sigma$)} & \multicolumn{1}{c|}{F1 ($\sigma$)} & \multicolumn{1}{c}{F1 ($\sigma$)} \\ \midrule
\multicolumn{5}{c}{Multilingual fine-tuned Models} \\
\midrule
\multicolumn{1}{l|}{\textbf{Multilingual BERT}} & \multicolumn{1}{c|}{59.81 (0.09)} & \multicolumn{1}{c|}{71.80 (0.12)} & \multicolumn{1}{c|}{80.85 (0.39)} & \multicolumn{1}{c}{91.64 (0.10)} \\ \midrule
\multicolumn{1}{l|}{\textbf{XLM-RoBERTa-base}}       & \multicolumn{1}{c|}{\textbf{76.57} (0.13)} & \multicolumn{1}{c|}{\textbf{82.69} (0.09)} & \multicolumn{1}{c|}{80.29 (0.53)} & \multicolumn{1}{c}{91.13 (0.16)} \\ \midrule
\multicolumn{5}{c}{Korean fine-tuned Models} \\ \midrule
\multicolumn{1}{l|}{\textbf{KLUE-BERT-base}}         & \multicolumn{1}{c|}{39.31 (0.10)} & \multicolumn{1}{c|}{55.63 (0.15)} & \multicolumn{1}{c|}{\textbf{82.44} (0.18)} & \multicolumn{1}{c}{\textbf{93.08} (0.27)} \\ \midrule
\multicolumn{1}{l|}{\textbf{KLUE-RoBERTa-base}}      & \multicolumn{1}{c|}{38.92 (0.28)} & \multicolumn{1}{c|}{55.47 (0.21)} & \multicolumn{1}{c|}{82.42 (0.57)} & \multicolumn{1}{c}{92.80 (0.17)} \\ \bottomrule[1.5pt]
\end{tabular}
}
\caption{Experiments results on the NER, RE, and ET tasks. F1 score (\%) is used for the evaluation metric with $\sigma$ which shows the standard deviation of the score. We divide the baseline models into two parts: the Multilingual models and the Korean models, marking the highest performances with bold text. \label{table:results}}
\end{table*}

\subsubsection{Statistics}
The overall statistics of \datasetname are showed in Table \ref{table:total-statistics}.
For the NER corpus, 457,232 entities from 112,362 examples in total.
For the RE corpus, 79,942 relations from 38,765 examples were annotated in total.
For the ET corpus, 332,830 entity mentions from 113,198 examples were annotated in total.
The annotated corpus was divided into three subsets for each task, i.e., a ratio of 8:1:1 for training, development, and testing, respectively. 
In this section, we describe our corpus statistically in the order of NER, RE, and ET. 

First, as shown in Table \ref{table:ner_label}, we used 12 entity types for our cultural heritage NER corpus.
Due to the properties of the cultural heritage domain, the three primary entity types, i.e., artifacts (AF), person (PS), and term (TM), account for the majority of the total entity population.
AF, PS, and TM entities possess approximately 36\%, 20\%, and 10\%, respectively, which are used as crucial information in the cultural heritage domain. The AF type includes cultural assets and historical landmarks, the TM type includes patterns or traces engraved on certain cultural assets, and the PS type particularly includes not only general people but also particular types of persons such as mythical figures.
On the other hand, the EV type occupies the most minor proportion, approximately 0.8\%, because our corpus especially aims to concentrate on the cultural heritage.

Second, Table \ref{table*:relabel} demonstrates the distribution of 14 RE labels.
In the case of ``A depicts B'' and ``A documents B,'' cultural assets left in a specific form such as records, drawings, and photographs are included, whereas ``A hasSection B" contains cultural heritage or historical landmarks located at a specific place.
Among them, ``A depicts B,'' ``A documents B,'' and ``A hasSection B'' are the most relationship labels with approximately 22\%, 16\%, and 10\% of the total, respectively.
``A depicts B'' and ``A documents B" include cultural assets left in a specific form such as records, drawings, and photographs, whereas ``A hasSection B" contains cultural heritage or historical landmarks located at a particular place.
``A hasDestroyed B" has the smallest proportion with ten relations in total because, in actual history, significant events such as the collapse of a nation or the loss of cultural properties are not as diverse as the types of general cultural assets.

Finally, among the fine-grained entity types, the ``AF\_DOCUMENTS" type, such as historical documents, occupies the largest part with 17.9\%, and ``PS\_NAME" including the names of historical figures, takes second place by occupying 11.5\%. On the other hand, the entity types to which belong to the AM, PT, MT, and EV almost account for under 1.0\%.

\section{Experiment}
The detailed experimental settings are in Appendix~\ref{appendix}.
\paragraph{Experimental results}
According to Table \ref{table:results}, two tendencies are observed.
One is that in the NER task, the multilingual models, i.e., multilingual BERT and xlm-RoBERTa-base, showed better performance by more than 30\% difference in both Entity F1 and Character F1 scores compared to the Korean models, i.e., KLUE-BERT-base and KLUE-RoBERTa-base.
The other is that in the RE and ET tasks, the performances of the Korean models were at least 1.1\% higher than those of the multilingual models.

\paragraph{Experimental Analysis}
As the token classification tasks are directly affected by segmentation~\citep{kim2021enhancing, park-etal-2021-find}, models with linguistic knowledge of Chinese and Japanese overperform in such tasks~\citep{pires2019multilingual}. In other words, the multilingual models are considered to segment better each token composed of various languages, especially in the NER corpus.
In addition, in Table \ref{table:ner-unk}, the Korean models, i.e., KLUE-BERT-base and KLUE-RoBERTa-base show a significantly higher ratio of unknown tokens than the multilingual language models.
It is attributed that the NER task requires more polyglot features of the model compared to the other tasks, i.e., RE and ET, which has the properties of sentence classification tasks.
On the other hand, as the RE or ET task does not classify all tokens in a sentence, the correct answer can be satisfactorily inferred from only the given Korean words; thereby, the language models pre-trained in Korean show better performance in the two tasks compared to the multilingual model.

\begin{table}[t]
\centering
\scalebox{0.775}{
\begin{tabular}{l|c|c}
\toprule[1.5pt]
\multicolumn{1}{c|}{\textbf{Model}} & \multicolumn{1}{c|}{\textbf{\begin{tabular}[c]{@{}c@{}}UNK\_dev (\%)\end{tabular}}} & \multicolumn{1}{c}{\textbf{\begin{tabular}[c]{@{}c@{}}UNK\_test (\%)\end{tabular}}} \\ \midrule
\textbf{Multilingual BERT}                       & 0.8156\%                                                                              & 0.7684\%                                                                               \\ \midrule
\textbf{XLM-RoBERTa-base}            & 0.1952\%                                                                              & 0.1810\%                                                                               \\ \midrule
\textbf{KLUE-BERT-base}              & 5.8670\%                                                                              & 5.9677\%                                                                               \\ \midrule
\textbf{KLUE-RoBERTa-base}           & 5.8670\%                                                                              & 5.9677\%                                                                               \\ \bottomrule[1.5pt]
\end{tabular}
}
\caption{\label{table:ner-unk}Unknown (UNK) token ratio (\%) of each model for development and testing set in the corpus. Baseline models pre-trained in Korean show the same proportions because they use identical vocabulary and tokenizers.}
\end{table}

\section{Conclusion}
In this paper, we introduced \textbf{\datasetname} - a Korean cultural heritage corpus for three typical entity-related tasks, i.e., NER, RE, and ET. 
Unlike the existing public Korean datasets with additional restrictions, \datasetname obviated the cumbersome prerequisite and can be freely modified and redistributed.
Furthermore, we proved the applicability of our entity-abundant corpus with the experiments employing the various pre-trained language models and provided practical insights regarding the statistical, diachronic, and linguistic analysis.
Above all, the most significant contributing point is that the disclosure of our corpus is expected to serve as a cornerstone for the development of IE tasks for a traditional cultural heritage. 
We hope that the continuous effort to preserve cultural heritage with the effective management of digitized documents containing cultural artifacts is encouraged by this research.

\section*{Acknowledgements}

This research is supported by Ministry of Culture, Sports and Tourism and Korea Creative Content Agency(Project Number: R2020040045), MSIT(Ministry of Science and ICT), Korea, under the ITRC(Information Technology Research Center) support program(IITP-2018-0-01405) supervised by the IITP(Institute for Information \& Communications Technology Planning \& Evaluation), and Institute of Information \& communications Technology Planning \& Evaluation(IITP) grant funded by the Korea government(MSIT) (No. 2020-0-00368, A Neural-Symbolic Model for Knowledge Acquisition and Inference Techniques).


\bibliographystyle{acl_natbib}
\bibliography{custom}

\appendix
\section{Experimental Setup}\label{appendix}

As the baseline models, we employed two global language models: multilingual bidirectional encoder representations from transformers (BERT)~\citep{devlin-etal-2019-bert} and a cross-lingual language model XLM-RoBERTa-base~\citep{conneau2020unsupervised} containing the Korean language, and two KLUE language models: KLUE-BERT-base, KLUE-RoBERTa-base, which were recently published covering various Korean downstream tasks.
In all the model experiments, the performance of each model was measured five times, and the average of each result was evaluated as the final result. Further, we set our environment for the experiment with four A6000 GPUs and 384 GB memory.
The hyperparameters in the fine-tuning step were set as follows.
The learning rate and weight decay were consistently set at 5e-5 and 0.01 across all three tasks.
The number of training epochs was set to 10 in NER, RE and 3 in ET. The batch size in training and testing procedures was set to 128 in NER, RE and 256 in ET. In the case of max sequence length, the lengths of 256 and 128 were used for each task.

We evaluated our system by employing F1 score, which is standard metric for classification tasks.
Specifically, the evaluation metrics for NER task were Entity F1 and Character F1 based on previous research~\citep{park2021klue}.
Entity F1 is a metric that is recognized as a correct answer only when all types included in an entity are matched accurately.
Conversely, Character F1 is a metric that evaluates each type of syllable in a sentence individually.
The evaluation metrics for the RE task were F1 score in the Scikit-learn library~\citep{JMLR:v12:pedregosa11a}.
As for ET, we adopted the evaluation metrics of loose F1 score following the same evaluation criteria used in previous works~\citep{ling2012fine, wang2020k}.

\end{document}